\documentclass[10pt,twocolumn,letterpaper]{article}

\usepackage{cvpr}      

\usepackage{caption,subcaption}
\usepackage{array}
\usepackage{amsmath}
\usepackage{times}
\usepackage{epsfig}
\usepackage{graphicx}
\usepackage{amsmath}
\usepackage{amssymb}
\usepackage{physics}
\usepackage{comment}

\usepackage{epsfig}
\usepackage{epstopdf}
\usepackage{booktabs}
\usepackage{algpseudocode} 
\usepackage{lipsum}
\usepackage{multirow}
\usepackage{textcomp}
\usepackage{gensymb}
\usepackage{pifont}
\usepackage{bm}

\usepackage{tabularx}
\usepackage{adjustbox}
\usepackage{array}
\usepackage{multirow}
\usepackage{url}
\usepackage{booktabs}
\usepackage{array}
\usepackage{lipsum}
\usepackage{makecell} 
\usepackage{caption, subcaption}
\usepackage{enumitem}

\usepackage[table]{xcolor}

\usepackage{float}
\definecolor{cvprblue}{rgb}{0.21,0.49,0.74}
\usepackage[pagebackref,breaklinks,colorlinks,citecolor=cvprblue]{hyperref}

\def\paperID{8780} 
\def\confName{CVPR}
\def\confYear{2024}

\title{S3Editor: A Sparse Semantic-Disentangled Self-Training \\ Framework for Face Video Editing}

\author{Guangzhi Wang$^{1}$, Tianyi Chen$^{2}$, Kamran Ghasedi$^{2}$, HsiangTao Wu$^{2}$, Tianyu Ding$^{2}$\\ 
Chris Nuesmeyer$^{2}$, Ilya Zharkov$^{2}$, Mohan Kankanhalli$^{1}$, Luming Liang$^{2}$\\
$^{1}$National University of Singapore, $^{2}$Microsoft\\
{\tt\small guangzhi.wang@u.nus.edu, mohan@comp.nus.edu.sg}\\{\tt\small \{tiachen,kamrang,musclewu,tianyuding,chnues,zharkov,lulian\}@microsoft.com}
}

\begin{document}
\twocolumn[{%
\maketitle
\begin{figure}[H]
    \hsize=\textwidth
    \centering
    \includegraphics[width=1.0\textwidth]{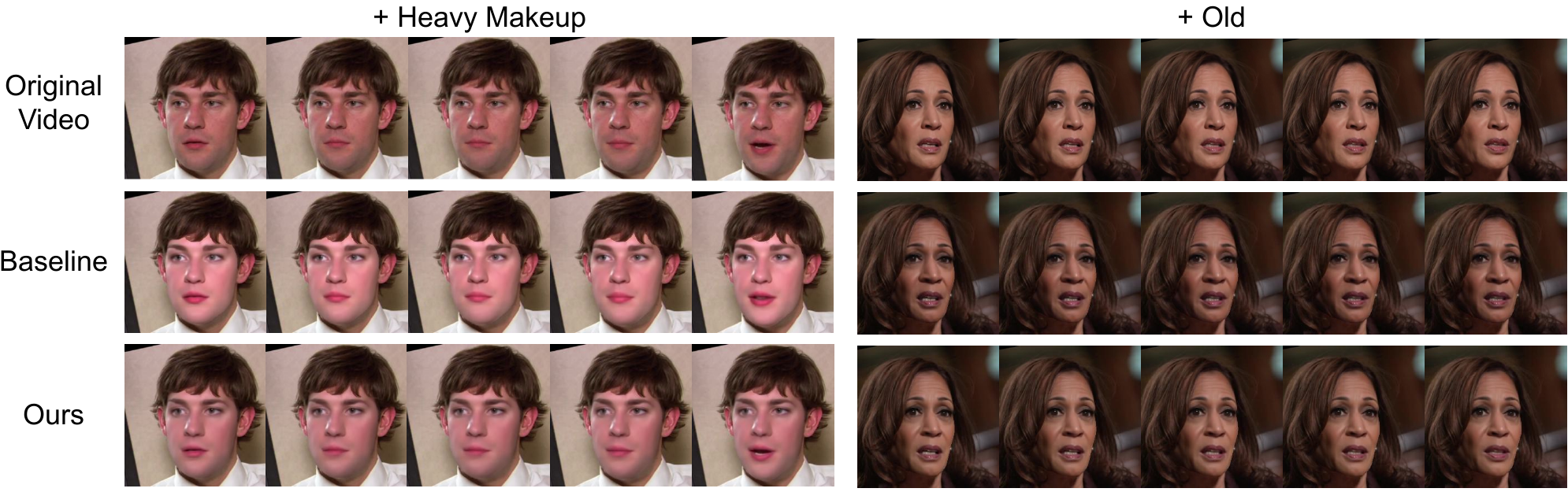}
    \caption{Face video editing results with diffusion video autoencoder~\cite{kim2023diffusion_videoae} and S3Editor (\textbf{S}parse \textbf{S}emantic-disentangled \textbf{S}elf-training) face video editing framework. S3Editor captures better editing faithfulness with better identity preservation (see the left), and avoid over-editing, \textit{e.g.,} the skin color on the right by the baseline method is unexpectedly affected by the hair color while ours preserve the skin color well. }
\end{figure}
}]

\newcommand{\algacro}{S3Editor{}}

\begin{abstract}
Face attribute editing plays a pivotal role in various applications. However, existing methods encounter challenges in achieving high-quality results while preserving identity, editing faithfulness, and temporal consistency.
These challenges are rooted in issues related to the training pipeline, including limited supervision, architecture design, and optimization strategy.
In this work, we introduce \algacro{}, a \textbf{S}parse \textbf{S}emantic-disentangled \textbf{S}elf-training framework for face video editing. \algacro{} is a generic solution that comprehensively addresses these challenges with three key contributions. Firstly, \algacro{} adopts a self-training paradigm to enhance the training process through semi-supervision. Secondly, we propose a semantic disentangled architecture with a dynamic routing mechanism that accommodates diverse editing requirements. Thirdly, we present a structured sparse optimization schema that identifies and deactivates malicious  neurons to further disentangle impacts from untarget attributes.
\algacro{} is model-agnostic and compatible with various editing approaches. Our extensive qualitative and quantitative results affirm that our approach significantly enhances identity preservation, editing fidelity, as well as temporal consistency.
\end{abstract}

\section{Introduction}\label{sec:intro}
Facial attribute editing~\cite{he2019attgan,zhu2020domain,yao2021latent,preechakul2022diffusionae,zhang2018generative, zhan2023multimodal} has facilitated extensive applications across diverse domains, capturing significant attention within the research community. In the scope of this research, our focus is on the intricate realm of face video editing~\cite{kim2023diffusion_videoae, yao2021latent, tzaban2022stitch, liu2022deepfacevideoediting}, an evolution from face image editing~\cite{zhu2020domain,shen2020interfacegan}. This task is notably more challenging as it strives to deliver not only high-quality results for each frame but also ensure temporal consistency among them, presenting a nuanced set of complexities.

	Early face video editing approaches~\cite{yao2021latent,tzaban2022stitch} predominantly exploit pre-trained StyleGAN~\cite{karras2019stylegan,Karras2020ada,shen2020interfacegan,ji2019generative} to streamline the editing process. In these methods, each frame undergoes an initial transformation into a representation within StyleGAN's latent space through an inversion process~\cite{richardson2021encoding, tov2021designing}, followed by the application of desired edits. However, the quality of editing is heavily contingent on the effectiveness of the GAN inversion process, serving as a bottleneck for overall results.
	More recently, diffusion models~\cite{song2020denoising_ddim, ho2020denoising_ddpm,nichol2021improved}, renowned for their strong generative capabilities, have demonstrated success in face image editing, outperforming GAN-based approaches in editing quality. In the case of diffusion autoencoders~\cite{preechakul2022diffusionae}, a U-Net~\cite{ronneberger2015unet,ding2022sparsity} is trained for image denoising, and an additional encoder is employed to encode facial features into a semantic representation, which can be further manipulated for the downstream editing. Building upon it, diffusion video autoencoders~\cite{kim2023diffusion_videoae} extend the paradigm to face video editing. In this framework, each frame is encoded by using a landmark detector~\cite{dong2018style,zhang2014facial} and an ArcFace~\cite{deng2019arcface}, serving as the condition that can be further transformed for editing purposes.

	{Recent years have seen notable progress in face video editing. However, current models still struggle to produce high-quality results while preserving identity, editing faithfulness, and temporal consistency.} 
These challenges stem from various constraints in the training pipeline: \textit{(i)} inadequate training supervision for desired editing outcomes, attributed to a scarcity of paired data, \textit{(ii)} suboptimal architecture lacking sufficient capacity to address diverse editing requirements, and \textit{(iii)} ineffective optimization strategy that excessively involves redundant neurons, leading to unintended alterations in regions that are supposed to remain unaffected. These limitations have impeded the refinement of face video editing methods, often overlooked in prior research. To tackle these issues, we introduce \algacro{}, an innovative (\textbf{S}parse \textbf{S}emantic-disentangled \textbf{S}elf-training) face video editing framework designed to comprehensively confront and overcome these challenges. \algacro{} encapsulates three key takeaways, summarized as follows.

\noindent\textbf{Self-Training for Generalizable Face Editing.} Within the domain of face video editing, a substantial obstacle lies in the scarcity of available supervised paired data. Existing methodologies~\cite{kim2023diffusion_videoae} often rely on training with face video datasets, such as VoxCeleb~\cite{nagrani2017voxceleb}, lacking the integration of explicit editing instructions or operations. Consequently, when edits are introduced during the inference stage, these methods may exhibit suboptimal performance, as their limited ability to generalize can give rise to artifacts and inconsistencies in identity, faithfulness, and temporality within the edited video.
In response to this challenge, we propose a self-training strategy aimed at achieving more robust and generalizable face video editing. Our approach initiates with a latent representation of a face, from which we generate pseudo-edited facial representations by uniformly sampling from an editing attribute pool. Subsequently, we meticulously design a set of objectives that encompass identity preservation and editing faithfulness, contributing to the semi-supervision of the training process. This self-training strategy significantly enhances the generalization capabilities of existing models, yielding superior editing results characterized by improved identity preservation, editing faithfulness, and enhanced temporal coherence.

\noindent
\textbf{Semantic Disentangled Editing Architecture.} Varied edits demand the encoding of facial features into distinct latent representations, emphasizing different facial regions. Acknowledging this need, we design a semantic disentangled architecture capable of catering to a diverse range of editing requirements. We classify all potential edits into multiple clusters upon their semantic representations, then establish a learnable transformation specific to each cluster. These transformations are dynamically activated based on the specific edit demand, contributing to an adaptive editing framework. The proposed semantic disentangled architecture significantly augments the model's capacity and effectively complements the introduced self-training strategies.

\noindent\textbf{Sparse Learning to Avoid Over-Editing.} Moreover, certain edits, such as \texttt{+bushy\_eyebrow}, necessitate precise modifications to localized regions of the original face while preserving the integrity of other areas. Notably, we have observed that existing methods often grapple with these scenarios, exhibiting a tendency to overly modify the entire face. To address this challenge, we introduce a tailored sparse learning strategy specifically designed for avoiding over-editing. This innovative approach involves partitioning facial latent representations into multiple distinct regions and actively promoting region sparsity during the training process. Through this refined strategy, the model learns to recognize and transform only the most pertinent facial areas for each specific edit. This results in a more precise editing process and also contributes to an overall enhancement of our semantic disentangled architecture.
Our proposed \algacro{} framework represents a substantial and cohesive advancement, accommodating various face video editing methods such as the GAN-based Latent Transformer \cite{yao2021latent} and the diffusion-based DiffVAE \cite{kim2023diffusion_videoae}. Comprehensive qualitative and quantitative results demonstrate that \algacro{} not only improves the editing faithfulness and identity preservation of individual frames, but also enhances temporal consistency and avoids over-editing. 

\section{Related Work}
\label{sec:formatting}

\begin{figure*}[t!]
	\centering
	\includegraphics[width=1.0\textwidth]{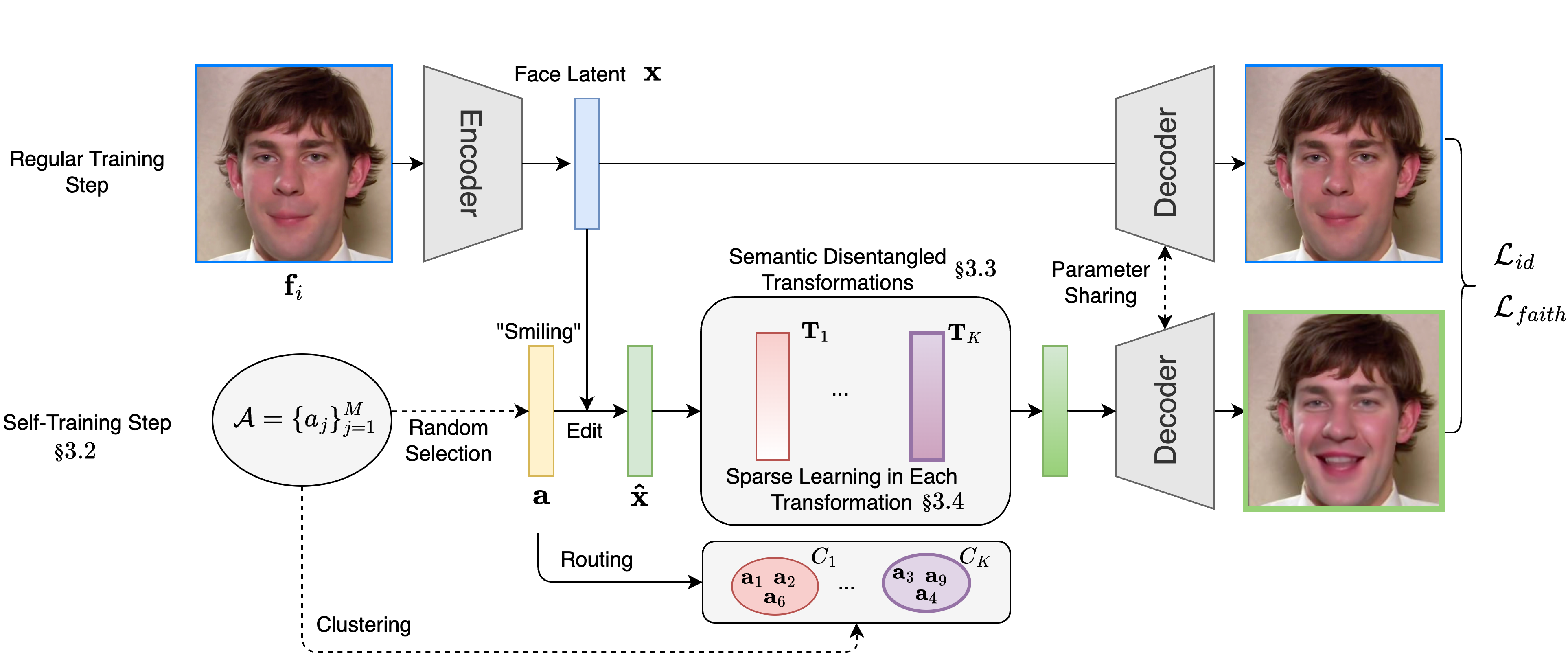}
	\caption{Overview of \algacro{} framework, which consists of three components to improve face video editing methods. \textit{(i)} Self-Training (Section~\ref{sec.self-training}): Given a face latent $\bm{x}$, we randomly select an attribute, \textit{e.g}, \texttt{Smiling}, to perform an edit operation on this latent, which is then taken as the condition for face generation. A set of optimization objectives using the edited face and the original face are designed to optimize the generation model. \textit{(ii)} Semantic disentangled editing architecture (Section~\ref{sec.arch}): We cluster all possible editing into $K$ clusters, and learn a set of transformations $\mathbf{T}_1 ... \mathbf{T}_K$ to conditionally encode the edited latent.
		\textit{(iii)} Sparse Learning to avoid over-editing (Section~\ref{sec.sparse_learning}): We encourage sparsity in each transformation to facilitate precise and localized editing.
	}\label{fig:framework}
\end{figure*}

\subsection{Face Image Editing}

The core concept in face editing revolves around manipulating the latent representation of a face, shifting it along a specific semantic direction~\cite{shen2020interfacegan}. This directional adjustment is typically derived from an attribute representation or natural language descriptions~\cite{parmar2023zero}. In the context of StyleGAN-based approaches, a pivotal phase in this manipulation is the GAN-Inversion process~\cite{richardson2021encoding, tov2021designing,zhu2020domain}. Here, the original face is encoded into a latent representation within StyleGAN's latent space to facilitate subsequent editing. Nevertheless, the intricacy of this process poses a significant challenge and currently acts as a bottleneck, impacting the overall editing quality of StyleGAN-based methods~\cite{yao2021latent}.

Recently, based on the strong generation capability of diffusion models~\cite{song2020denoising_ddim, ho2020denoising_ddpm, dhariwal2021diffusion_beatsgan,zhou2023dream}, recent work also utilized it for image editing~\cite{mokady2023null, hertz2022prompt, zhang2023text}.
For face image editing, besides encoding the image into a noisy map with a U-Net~\cite{ronneberger2015unet} architecture, diffusion autoencoder~\cite{preechakul2022diffusionae,rey2019diffusion} also encodes the face representation with an extra network, which contains the semantic representation of the face and can be further manipulated for face editing.

In attribute-based face editing, the attributes are predominantly sourced from an attribute classifier that is pre-trained on ArcFace's latent space. To broaden the scope of applicability for face editing techniques, recent advancements, such as StyleClip~\cite{patashnik2021styleclip}, have introduced the use of natural language prompts~\cite{brooks2023instructpix2pix}. This innovative approach involves leveraging CLIP~\cite{radford2021learning_clip}, which is pre-trained on extensive image-text pairs. By utilizing this methodology, the semantic editing direction is determined by the disparity between the representations of two prompts—one with and the other without related attributes. 

\subsection{Face Video Editing}
In contrast to face image editing, face video editing poses more challenges, since it necessitates not only meticulous editing of each frame but also the preservation of temporal consistency across frames. Early approaches employed StyleGAN~\cite{yao2021latent} and the Crop \& Align strategy~\cite{tzaban2022stitch} to address this challenge, yet encountered shortcomings in maintaining sufficient temporal consistency. Recent advancements, leveraging the decoupled semantic representation of faces in diffusion autoencoder~\cite{preechakul2022diffusionae}, have yielded notable progress. In particular, the diffusion video autoencoder~\cite{kim2023diffusion_videoae} encodes the face into two disentangled semantic representations: one for identity and another for landmark features. The former facilitates semantic face editing, while the latter preserves motion information. By retaining facial motion details within the landmark representations, the original video's temporal consistency is significantly improved. In addition, there exist  methods~\cite{ceylan2023pix2video,yang2023rerender,chai2023stablevideo,geyer2023tokenflow,molad2023dreamix,singer2022make,yu2022generating, yang2023styleganex, blattmann2023align} that are designed for general video editing could apply onto face video editing as well, though are typically not competitive to the domain-expertise methods.

\section{Methodology}

\subsection{Overview}

	Face video editing endeavors to effectively manipulate the original video while preserving both the facial identity and temporal consistency. We specifically delve into the domain of attribute-based human-face video editing. Formally, given a human face video $\mathcal{V} = \{\bm{f}_1, \bm{f}_2, ..., \bm{f}_N\}$ comprised of $N$ sequential frames, our objective is to manipulate this video with a designated attribute $a \in \mathcal{A}$, where $\mathcal{A} = \{a_j\}_{j=1}^{M}$ represents the set of all conceivable edits. For semantic attribute editing, each attribute $a_j \in \mathcal{A}$ is embedded into a semantic representation $\mathbf{a}_j\in\mathbb{R}^d$. 

	Prior efforts in face video editing have explored the use of StyleGAN~\cite{karras2019stylegan,Karras2021} and diffusion models~\cite{preechakul2022diffusionae, kim2023diffusion_videoae}. The underlying editing model can be succinctly represented as $\mathcal{M} = \{\mathcal{E}, \mathcal{D} \}$, where the former, $\mathcal{E}$, denotes an encoder which encodes a face into its latent representation, while the latter, $\mathcal{D}$, represents a decoder generating a face from the latent representation. Despite these advancements, current methods continue to grapple with challenges in producing high-quality results. We posit that these challenges arise from issues within the training pipeline, specifically: \textit{(i)} restricted training supervision due to the scarcity of paired data, \textit{(ii)} suboptimal architectural design insufficient for addressing diverse editing requests, and \textit{(iii)} ineffective training strategies leading to over-editing, \ie, distorted regions that should not be affected given the target editing.
	
	In response to these challenges, we introduce a novel  (\textbf{S}parse \textbf{S}emantic-disentangled \textbf{S}elf-training) face video editing framework—\algacro{}. \algacro{} is meticulously crafted to comprehensively tackle these three key issues and is adaptable across various editing methods. It encompasses: \textit{(i)} a self-training strategy that semi-supervises training for more generalizable face editing, \textit{(ii)} a disentangled architecture adept at handling diverse editing requirements, and \textit{(iii)} a sparsity-driven learning framework facilitating precise, localized editing. The entire pipeline is depicted in Figure~\ref{fig:framework}, and each of these innovative components will be elaborated in the following subsections.

\subsection{Self-Training for Generalizable Face Editing}\label{sec.self-training}

	Existing approaches predominantly rely on generative models, such as StyleGAN~\cite{karras2019stylegan} and diffusion models~\cite{kim2023diffusion_videoae}, to execute face editing. The editing process occurs on the latent representation of the original face, which afterward serves as the condition for the generative model to generate an edited face. However, these generative models are typically trained on face video datasets without explicit editing signals during training. Consequently, when deployed for face editing, the pre-trained generation models encounter difficulties in achieving high-quality and coherent results, primarily due to limited generalizability.
	A straightforward resolution to this challenge may involve collecting video data with editing annotations for joint training of the generative model. However, this is often cost-prohibitive and impractical, particularly for imaginative edits such as gender transformation. In response to these limitations, our work introduces an innovative self-training strategy explicitly designed to enhance the generalizability of generative models for diverse face editing tasks. 

	Given a frame $\bm{f}\in \mathcal{V}$\footnote{We omit the subscript for simplicity.}, it is encoded into the latent space with the encoder, resulting in its representation $\bm{x} \gets \mathcal{E}(\mathbf{f})$. In the conventional training process, the face representation $\bm{x}$ acts as the condition for the generation model to produce face images. This study introduces a novel approach to enhance model generalizability during training. Specifically, given $\bm{x}$, we uniformly randomly select an attribute $a\in \mathcal{A}$ with an embedded representation $\mathbf{a}\in\mathbb{R}^d$. Then we perform a pseudo-edit step on this latent representation as follows.
	\begin{equation}
		\bm{\hat{x}} \gets \mathbf{T}(\text{Denormalize}(\text{Normalize}(\bm{x}) + \gamma \cdot \bm{a})),
	\end{equation}
	where $\gamma$ is a randomly selected edit scale $\gamma \sim \text{Uniform}(0, 1)$ to control the aggressiveness of editing affects. $\mathbf{T}(\cdot)$ represents a learnable transformation to encode the edited latent into the original latent space. Subsequently, the edited latent $\bm{\hat{x}}$ is utilized as the condition to generate an edited face with the Decoder $\bm{\hat{f}} \gets \mathcal{D}(\bm{\hat{x}})$. Such pseudo-edit process produces numerous paired-wise training samples that require dedicately designed objective functions to effectively capture the editing faithfulness.
	
	To effectively learn from the semantics in the pseudo-edited latent $\bm{\hat{x}}$, we expect the generated face to maintain the original identity and be faithful to the selected edit $a\in \mathcal{A}$. As such, we design the following objectives to train the generative model:
	
	\begin{align}
		\mathcal{L}_\text{overall}:=&  \lambda_\text{id} \mathcal{L}_{id} + \lambda_\text{faith} \mathcal{L}_\text{faith} + \lambda_\text{gen} \mathcal{L}_\text{gen}\label{loss:overall}\\
		\mathcal{L}_\text{id}:=&  \norm{\text{Arcface}(\bm{f}) - \text{Arcface}(\bm{\hat{f}})} \\
		\mathcal{L}_\text{faith}:= & \sum_{\substack{a'\in \mathcal{A}\\a'\neq a}} \norm{[\text{Attr}(\bm{f})]_{a'}-[\text{Attr}(\bm{\hat{f}})]_{a'}} \nonumber\\
		&-\gamma\norm{[\text{Attr}(\bm{f})]_{a}-[\text{Attr}(\bm{\hat{f}})]_{a}}
	\end{align}
	Here, Arcface($\cdot$) refers to the pre-trained model~\cite{deng2019arcface} extracting the identity representation of a certain frame. The identity loss, $\mathcal{L}_\text{id}$, aims to encourage the pseudo-edited frame retains the original identity post-editing. Attr($\cdot$) refers to the pre-trained face attribute classifier producing the logits of each attribute in $\mathcal{A}$. The fidelity loss, $\mathcal{L}_\text{faith}$, encourages the edited frame $\bm{\hat{f}}$ to align accurately with the chosen attribute ${a}\in\mathcal{A}$ while preserving the integrity of other attributes that are not selected. Furthermore, $\mathcal{L}_\text{gen}$ represents the standard generation loss, a common objective in original generative models such as the denoising loss~\cite{kim2023diffusion_videoae}. The hyperparameters $\lambda_\text{id}$, $\lambda_\text{faith}$, and $\lambda_\text{gen}$ are employed to balance the influence of these three distinct objectives: identity preservation, attribute fidelity, and generation, respectively.
	
	
	\subsection{Semantic Disentangled Editing Architecture}\label{sec.arch}
	
	Existing face video editing methods typically transform a face frame $\bm{f}$ into a latent representation $\bm{x}$, which then undergoes a \textit{single} transformation before being input to a decoder $\mathcal{D}$ for content generation. However, this one-size-fits-all paradigm may not adequately address the diverse requirements of various editing tasks. For example, editing the eyes might necessitate a different transformation on the face frame compared to editing the mouth. This variability, or editing-level heterogeneity, challenges the efficacy of using a single transformation. Meanwhile, our self-training strategy, as detailed in Section~\ref{sec.self-training}, generates numerous pseudo-edited samples during the training process. Effectively managing these heterogeneous samples is crucial for the success of our approach.
	
	To overcome this challenge, we propose a semantic disentangled editing architecture. This architecture dynamically activates its processing route based on the specific facial attribute being edited, effectively addressing instance-level heterogeneity and providing a more tailored and precise editing outcome. Specifically, we begin by considering a set of editable attributes, denoted as $\mathcal{A}$. These attributes have their respective embedded representations $\{\mathbf{a}_j\}_{j=1}^M$. We apply the K-Means clustering algorithm \cite{arthur2006slow} to group these representations into $K$ disjoint clusters ${\mathcal{C}_1, \mathcal{C}_2, \cdots, \mathcal{C}_K}$. This grouping is based on semantic similarity, ensuring that attributes or edits that are similar over semantic representations are placed in the same cluster.
	For each cluster $\mathcal{C}_k$, we assign a specific transformation $\mathbf{T}_k$. This transformation is applied to the latent representation of the face and is dynamically chosen based on the particular edit being applied. Our disentangled architecture design offers several advantages. It \textit{(i)} enables the model to more effectively handle a variety of editing tasks and \textit{(ii)} enhances the utility of our proposed self-training process, leading to more versatile and generalizable editing capabilities.
	
	We would like to emphasize that our proposed semantic disentangled editing architecture significantly differs from the disentangled editing transformation in LatentTransformer~\cite{yao2021latent}. LatentTransformer assigns a unique transformation to each attribute, which has limitations, especially when encountering new, unseen attributes and in terms of the resources required to maintain a large number of attribute-specific transformations. Our method overcomes these challenges with two key advantages:
	\textit{(i) Generality}: Our architecture is more adaptable to new attributes by activating a module with the closest semantic similarity to the new attribute, rather than necessitating a predefined transformation for each possible attribute.
	\textit{(ii) Resource Efficiency}: Our approach is significantly more resource-efficient. Instead of scaling the number of transformations with the number of attributes, we maintain a limited number of cluster transformations. This design not only conserves computational resources but also simplifies the model's scalability, making it more practical for a wider range of applications.

	\subsection{Sparse Learning to Avoid Over-Editing}\label{sec.sparse_learning}
	
	\begin{figure}
		\vspace{2mm}
		\centering
		\includegraphics[width=1.0\columnwidth]{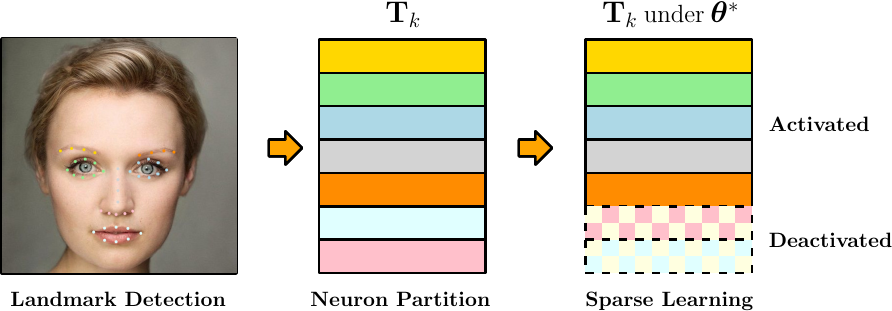}
		\caption{Sparse learning for localized editing. Facial landmarks extracted by a pre-trained detector are clustered according to the geometrical proximity. We encourage structured sparsity in each transformation $\mathbf{T}_k$ to identify and deactivate malicious neurons to facilitate the editing precision.}
		\label{fig:sparse_learning}
	\end{figure}
	
		A significant challenge in current facial editing methodologies is the issue of over-editing, \ie, transforming the areas that should be not changed. There are scenarios where a comprehensive, global adjustment to an image is desired. However, for certain edits, such as adding \texttt{Bushy\_Eyebrows}, affecting only a specific portion of the image is preferred. Previous studies have often overlooked this dichotomy, leading to suboptimal editing outcomes. We believe this shortfall arises because the activations of neurons within the editing model are intertwined; some neurons inadvertently have a detrimental influence on the outcome. To circumvent this, we introduce a group sparse learning technique into the self-training strategy upon the semantic disentangled architecture
		aimed specifically at refining the granularity of \textit{localized edits}, ensuring that modifications are made precisely where intended. 
	
		\paragraph{Neuron Partition.} Within the editing model $\mathcal{M}$, given the trainable neurons denoted as $\bm{\theta}$, the initial step involves segregating the neurons into distinct groups $\mathcal{G}$. This partitioning is adaptable, and subject to various criteria. For instance, in methodologies like Diffusion Video Autoencoder~\cite{kim2023diffusion_videoae}, a face is intuitively segmented into multiple sections based on its facial landmark representations as Figure~\ref{fig:sparse_learning}.
		Subsequently, each facial segment is transmuted into a latent representation through a designated transformation. Our approach favors a sparse implementation of this transformation; that is, certain transformations are selectively deactivated. As a result, during post-editing, specific portions of the face representation remain unaltered and preserve the original features in those regions. The definition of a group hinges on the transformation executed upon it, ensuring that only the targeted areas are modified.
	
		\paragraph{Structured Sparse Learning.} Upon the neuron partition $\mathcal{G}$, we then employ structured sparsity learning to identify the neurons to be deactivated. Such a problem can be formulated as follows.
		\begin{equation}\label{main:prob}
			\begin{split}
				\mathop{\text{minimize}}_{\bm{\theta}\in\mathbb{R}^n}&\ \mathcal{L}_\text{overall}\\
				\text{s.t.}&\ \text{Cardinality}\{g|g\in\mathcal{G}\ \text{and} \ [\bm{\theta}]_g=\bm{0}\}=Q,
			\end{split}
		\end{equation}
		where $Q$ is the target sparsity level, and the cardinality measures the size of zero groups in $\mathcal{G}$.
		To solve problem~\eqref{main:prob}, we build upon the state-of-art sparse optimizer DHSPG~\cite{chen2021only,chen2023otov2,dai2023adaptive} and stabilize its sparsity exploration. In particular, DHSPG at first converts the sparsity constraints in problem \eqref{main:prob} into an explicit regularization term to form an unconstrained optimization problem as follows. After a warm-up phase, a subset of groups of neurons are then progressively pushed to the origin, and projected as zero to be deactivated. 
		\begin{equation}
			\mathop{\text{minimize}}_{\bm{\theta}\in\mathbb{R}^n}\ \mathcal{L}_\text{overall} +\sum_{g\in\mathcal{G}}\lambda_g \norm{[\bm{\theta}]_g}_2,
		\end{equation}
		where $\lambda_g$ is the regularization coefficient for each $g\in\mathcal{G}$. The larger coefficient typically yields sparsity more aggressively. We found its selection depends on time-consuming hyper-parameter tuning efforts. Such inconvenience further results in unreliable sparsity exploration, \ie, the final solution may not reach the target sparsity level $Q$ under improperly selected $\lambda_g$. To mitigate the issues, we propose an implicit regularization schema. In general, after updating a trial iterate based on some first-order optimization, we automatically compute $\lambda_g$ for each $g\in\mathcal{G}$ such that the magnitude of $[\bm{\theta}]_g$ must be reduced at least by some certain degree. The computation of $\lambda_g$ is performed upon a line-search schema, where we start from an initial value, and increase it till the magnitude reduction meets a desired level. As a result, our enhanced optimizer is less sensitive to the hyper-parameter selection compared to the basic DHSPG. The group sparse iterate with the best evaluation performance $\bm{\theta}^*$ is returned for further editing usage. 
		
\begin{figure*}
	\centering
	\includegraphics[width=1.0\textwidth]{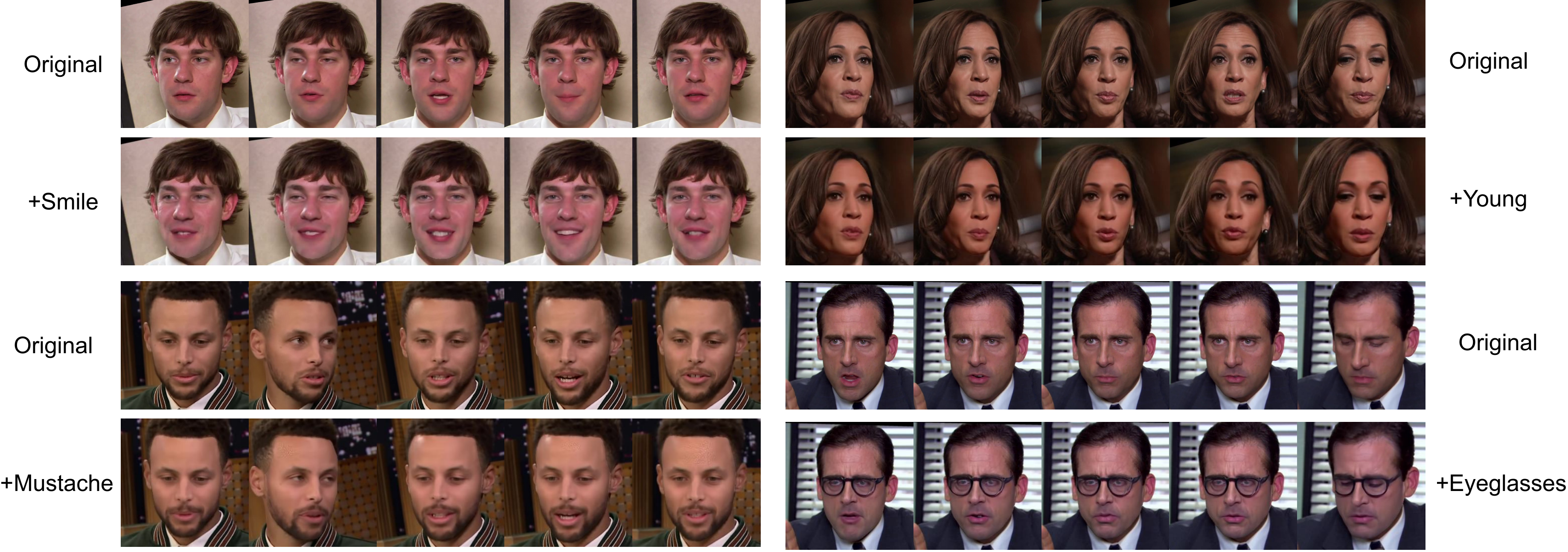}
	\caption{Qualitative results produced by \algacro{}. \algacro{} is able to preserve temporal consistency for motion-intensive edit (\texttt{+Smile}) and maintain original identity (\texttt{+Young}) while maintaining editing locality (\texttt{+Mustache}, \texttt{+Eyeglasses}).  }\label{fig:qualit}
\end{figure*}

\section{Experiments}

Our proposed \algacro{} framework represents a significant advancement over current editing techniques. It is \textit{generic} to integrate with a variety of editing methods, thereby enabling more generalized, temporally consistent, and high-quality face video editing. In this work, we demonstrate the application of \algacro{} through two notable face video editing methods: the Diffusion Video Autoencoder~\cite{kim2023diffusion_videoae} and the Latent Transformer~\cite{yao2021latent}.

\noindent\textbf{Diffusion Video Autoencoder (DiffVAE)~\cite{kim2023diffusion_videoae}} stands as the leading method in face video editing, utilizing advanced diffusion models. It employs a separate representation approach, separating the facial identity and landmarks into distinct features. To align with \algacro{}, within the architecture of DiffVAE, two dedicated encoders and a U-Net are responsible for the encoding process $\mathcal{E}$, while the U-Net is also utilized for denoising, functioning as the decoder $\mathcal{D}$. As outlined by~\cite{kim2023diffusion_videoae}, editing focuses exclusively on the facial identity aspect. To implement a sparse learning strategy, we categorize the neurons that transform the 51-point landmark points into seven neuron groups, aligning them with specific landmark positions, \eg, nose, eye, mouth, etc. This categorization fosters group sparsity based on the landmarks, and deactivates the malicious neurons, thereby enhancing the precision of the editing outcomes.

	
	\noindent\textbf{Latent Transformer~\cite{yao2021latent}} is a typical GAN-based face editing method that manipulates the faces with the usage of the GAN-Inversion~\cite{zhu2020domain,xia2022gan} process. To integrate this method within \algacro{} framework, we utilize the GAN-Inversion process as the encoder $\mathcal{E}$ and the GAN generator as the decoder $\mathcal{D}$. For implementing sparse learning, we treat each row of the transformation matrix as a distinct group. Subsequently, we apply sparsity regularization across these groups to avoid over-editing.

\subsection{Implementation Details}
\algacro{} framework is able to accommodate the aforementioned two methods.
For both baseline methods, we follow their default training protocol. For the self-training step, we set the values of balancing coefficients $\lambda_\text{id}$, $\lambda_\text{faith}$ $\lambda_\text{gen}$ as 0.5, 0.1 and 1, respectively. The number of semantic disentangled modules $K$ is set to 5. The group sparsity rate $Q$ is fixed to $\lfloor 0.1|\mathcal{G}|\rfloor$ to filter out 10\% malicious neurons.
Following previous work~\cite{kim2023diffusion_videoae}, given their pre-trained checkpoints, we further tune them with \algacro{} framework using the videos in the VoxCeleb dataset~\cite{nagrani2017voxceleb}, whereby each frame is resized to 256$\times$256 resolution. More implementation details could be found in the supplementary material. The source code will be available to the public.

\subsubsection{Qualitative Evaluation}
We provide qualitative results produced by \algacro{} in Figure~\ref{fig:qualit}.
The results show that \algacro{} successfully captures editing faithfulness, preserves identity, maintains temporal consistency, and avoids over-editing. For example, in the top-left of Figure~\ref{fig:qualit}, we edit the video with \texttt{+Smile}, which is challenging and requires motion variation. Along with the top-right, we observe that \algacro{} produces high-quality editing results with strong temporal consistency.
For the bottom-left, the video is manipulated with \texttt{+Mustache}. In the bottom-right, \algacro{} produces localized editing, which only modifies the glass areas while keeping other areas unchanged.

\begin{table}[]
	\centering
	\caption{Quantitative comparison before and after using \algacro{} framework on two baselines.}\label{tab:compare_sota}
	\scalebox{0.75}{
		\begin{tabular}{lccccc}
			\toprule
			Method & TL-ID & TG-ID & ID-Preserve & TACR & NAPR \\
			\midrule
			Stitch~\cite{tzaban2022stitch} &  0.996 & 0.991 & 0.8539 & 0.015 & 0.911 \\
			TCSVE~\cite{xu2022temporally}  &  0.995 & 0.990 & 0.9946 & 0.025 & 0.923 \\
			\midrule
			LatentTransformer~\cite{yao2021latent} & 0.9846 & 0.9813 & 0.9231 & 0.043 & 0.8973 \\
			+ Ours & \cellcolor{red!40}{0.9891} & \cellcolor{red!40}{0.9851} & \cellcolor{red!40}{0.9362} & \cellcolor{red!40}{0.066} & \cellcolor{red!40}{0.9152} \\
			\midrule
			DiffVAE~\cite{kim2023diffusion_videoae} & 0.9941 & 0.9923 & 0.9169 & 0.046 & 0.9460\\
			+ Ours & \cellcolor{red!40}{0.9986} & \cellcolor{red!40}{0.9972} & \cellcolor{red!40}{0.9456} & \cellcolor{red!40}{0.066} & \cellcolor{red!40}{0.9561} \\
			\bottomrule
		\end{tabular}
	}
\end{table}

\begin{table}[]
	\centering
	\caption{Effectiveness of each component in our method.}\label{tab:ablation_components}
	\scalebox{0.8}{
		\begin{tabular}{lccccc}
			\toprule
			Method & TL-ID & TG-ID & ID-Preserve & TACR & NAPR \\
			\midrule
			DiffVAE~\cite{kim2023diffusion_videoae} & 0.9941 & 0.9923 & 0.9169 & 0.046 & 0.9460 \\
			+ST & \cellcolor{orange!40}0.9979 & \cellcolor{orange!40}0.9971 & \cellcolor{yellow!40}0.9150 & \cellcolor{yellow!40}0.050 & \cellcolor{yellow!40}0.9548 \\
			+ST+SDA  & \cellcolor{yellow!40}0.9971 & \cellcolor{yellow!40}0.9962 & \cellcolor{orange!40}0.9436 & \cellcolor{orange!40}0.062 & \cellcolor{orange!40}0.9558 \\
			+ST+SDA+SL  & \cellcolor{red!40}0.9986 & \cellcolor{red!40}0.9972 & \cellcolor{red!40}0.9456 & \cellcolor{red!40}0.066 & \cellcolor{red!40}0.9561 \\
			\bottomrule
		\end{tabular}
	}
\end{table}

\begin{table}[]
	\centering
	\caption{Effectiveness on the number of clusters.}\label{tab:ablation_cluster}
	\scalebox{0.8}{
		\begin{tabular}{lccccc}
			\toprule
			\# Cluster & TL-ID & TG-ID & ID-Preserve & TACR & NAPR \\
			\midrule
			DiffVAE~\cite{kim2023diffusion_videoae} & 0.9941 & 0.9923 & 0.9169 & 0.046 & 0.9460 \\
			\midrule
			2  & 0.9944 & 0.9936 & 0.9256 & 0.052 & 0.9482 \\
			3  & \cellcolor{yellow!40}0.9956 & \cellcolor{orange!40}0.9943 & \cellcolor{yellow!40}0.9361 & \cellcolor{orange!40}0.058 & \cellcolor{red!40}0.9588 \\
			5  & \cellcolor{orange!40}0.9971 & \cellcolor{red!40}0.9962 & \cellcolor{red!40}0.9436 & \cellcolor{red!40}0.062 & \cellcolor{yellow!40}0.9558 \\
			10 & \cellcolor{red!40}0.9972 & \cellcolor{red!40}0.9962 & \cellcolor{orange!40}0.9421 & \cellcolor{orange!40}0.058 & \cellcolor{orange!40}0.9562 \\
			\bottomrule
		\end{tabular}
	}
\end{table}

\subsubsection{Quantitative Evaluation}

For quantitative evaluation, we compared the evaluation metrics that have been adopted in previous work, including \textit{(i)} editing faithfulness, \textit{(ii)} identity preservation, and \textit{(iii)} temporal consistency.

\noindent\textbf{Editing Faithfulness} quantifies how well the editing is performed with the given model. 
It consists of two parts: Target Attribute Change Rate (TACR) and Non-target Attribute Preservation Rate (NAPR).
The former measures the percentage of frames that the target attribute has been changed, while the latter computes the frame percentage where the non-target attributes have been preserved.

\noindent\textbf{Identity Preservation} computes the identity similarity between the edited frame and the original frame and is averaged across all frames.

\noindent\textbf{Temporal Consistency} measures relative temporal coherency between the edited video and the original video.
We mainly adopt two metrics: TG-ID and TL-ID~\cite{tzaban2022stitch}. 
TG-ID measures the identity similarity for all frame pairs in the edited video, normalized by the corresponding similarity in the original video.
In contrast, TL-ID considers and computes the similarity for every two adjacent frames only.

We show the comparison of these metrics in
Table~\ref{tab:compare_sota}. 
It can be observed that \algacro{} framework is able to consistently improve both baseline methods while surpassing previous benchmarks~\cite{tzaban2022stitch, xu2022temporally}.
Notably,  although \textit{no} temporal consistency specific constraints have been applied in our framework, the temporal consistency has been still enhanced upon both baseline models.
Such bonus may be due to generalizability enhanced by \algacro{}, which makes the model better handle various editing scenarios. 
Besides, our method can also improve identity preservation and editing faithfulness, with uniformly improved quantitative results on ID-Preserve, TACR, and NAPR.

\begin{table}[]
	\centering
	\caption{Effectiveness of Group Sparsity Rates.}\label{tab:ablation_group_sparsity}
	\scalebox{0.8}{
		\begin{tabular}{lccccc}
			\toprule
			Sparsity Rates & TL-ID & TG-ID & ID-Preserve & TACR & NAPR \\
			\midrule
			0.0 & \cellcolor{orange!40}0.9971 & \cellcolor{orange!40}0.9962 & \cellcolor{orange!40}0.9436 & \cellcolor{orange!40}0.062 & \cellcolor{orange!40}0.9558\\
			0.1  & \cellcolor{red!40}0.9986 & \cellcolor{red!40}0.9972 & \cellcolor{red!40}0.9456 & \cellcolor{red!40}0.066 & \cellcolor{red!40}0.9561 \\
			0.3  & \cellcolor{yellow!40}0.9234 & \cellcolor{yellow!40}0.9221 & \cellcolor{yellow!40}0.9120 & \cellcolor{orange!40}0.062 & \cellcolor{yellow!40}0.9321 \\
			0.5  & 0.8721 & 0.8624 & 0.9020 & 0.056 & 0.9124 \\
			\bottomrule
		\end{tabular}
	}
\end{table}

\begin{table}[]
	\centering
	\caption{Generalizaibility to unseen edits.}\label{tab:ablation_unseen}
	\vspace{-1em}
	\scalebox{0.8}{
		\begin{tabular}{l|ccccc}
			\toprule
			Method & TL-ID & TG-ID & ID-Preserve & TACR & NAPR   \\
			\midrule
			& \multicolumn{5}{c}{\textit{Seen Edits}} \\
			DiffVAE~\cite{kim2023diffusion_videoae} & 0.9935 & 0.9920 & 0.9069 & 0.044 & 0.9250 \\  
			+ Ours & \cellcolor{red!40}{0.9942} & \cellcolor{red!40}{0.9950} & \cellcolor{red!40}{0.9284} & \cellcolor{red!40}{0.050} & \cellcolor{red!40}{0.9362} 
			\\
			\midrule
			& \multicolumn{5}{c}{\textit{Unseen Edits}} \\
			DiffVAE~\cite{kim2023diffusion_videoae} & 0.9944 & 0.9928 & 0.9199 & 0.051 & 0.9464 \\
			+ Ours & \cellcolor{red!40}{0.9948} & \cellcolor{red!40}{0.9932} & \cellcolor{red!40}{0.9212} & \cellcolor{red!40}{0.056} & \cellcolor{red!40}{0.9488} \\
			\bottomrule
		\end{tabular}
	}
\end{table}

\subsection{Ablation Study}
We ablatively study the design of each component in our method. 
For all the experiments in this subsection, we take Diffusion Video Autoencoder~\cite{kim2023diffusion_videoae} as the baseline and use the same videos as in Table~\ref{tab:compare_sota} for comparison.

\noindent\textbf{Effectiveness of each component in \algacro{}.}
We study the effectiveness of each component in our proposed method, including the Self-Training strategy (ST), The design of our Semantic Disentangled Architecture (SDA), and the Sparse Learning (SL) strategy.
The comparison is shown in Table~\ref{tab:ablation_components}.
It can be observed that, with ST, our method can significantly improve the temporal consistency (TL-ID, TG-ID) of the baseline, even though no temporal constraints are applied.
Furthermore, with the incorporation of the SDA, we also observe that all ID-Preserve, TACR, and NAPR get further improved.
These results indicate that the original unified architecture with a single transformation bottlenecks its editing quality, making it struggle to handle various editing scenarios.
In contrast, our disentangled architecture makes it more performant by introducing more model capacity with limited cost.
Finally, with our Sparse Learning (SL) strategy for localized editing, we can further improve temporal consistency and ID preservation.
This can be related to the effect of our method that can minimally manipulate the original video when performing the editing, which leads to better maintenance of the contents in the original video while being faithful to the editing.

\noindent\textbf{The Number of Clusters.}
We study the number of clusters in SDA in Table~\ref{tab:ablation_cluster}. With only 2 clusters, our method has already improved the baseline across all metrics. With 5 clusters, our method achieves the best quantitative results.
We observe that, with 10 clusters, the performance almost saturates, indicating using more complex architecture does not always lead to performance improvements.

\noindent\textbf{Generalization to Unseen Editing.}
Our semantic disentangled architecture requires us to have a codebook of all editable attributes at hand for the clustering operation.
We study if our method is still generalizable to unseen attributes/edits under the current framework.
In particular, we split all 40 editable attributes into two groups: 30 attributes are available for training, and the rest 10 are kept unseen for training and used only for validation. 
The results are shown in Table~\ref{tab:ablation_unseen}.
It can be observed that our method improves the baseline's performance on editing quality on both seen and unseen attributes, demonstrating the strong generalization potential of our method.

\noindent\textbf{Sparsity Rates.}
To understand how many neurons are malicious for the editing outcomes, we perform an ablation study in Table~\ref{tab:ablation_group_sparsity} by iterating target sparsity level in $\{0.0, 0.1, 0.3, 0.5\}$. 
We observe that a sparsity rate of 0.1 gives the best results while further increasing this value leads to a severe performance degradation. This phenomenon indicates that most of the face areas are useful for face editing while a small portion of them would be useless for some certain edits. Deactivating the corresponding neurons could lead significant performance improvement.

\begin{figure}
	\includegraphics[width=1.0\columnwidth]{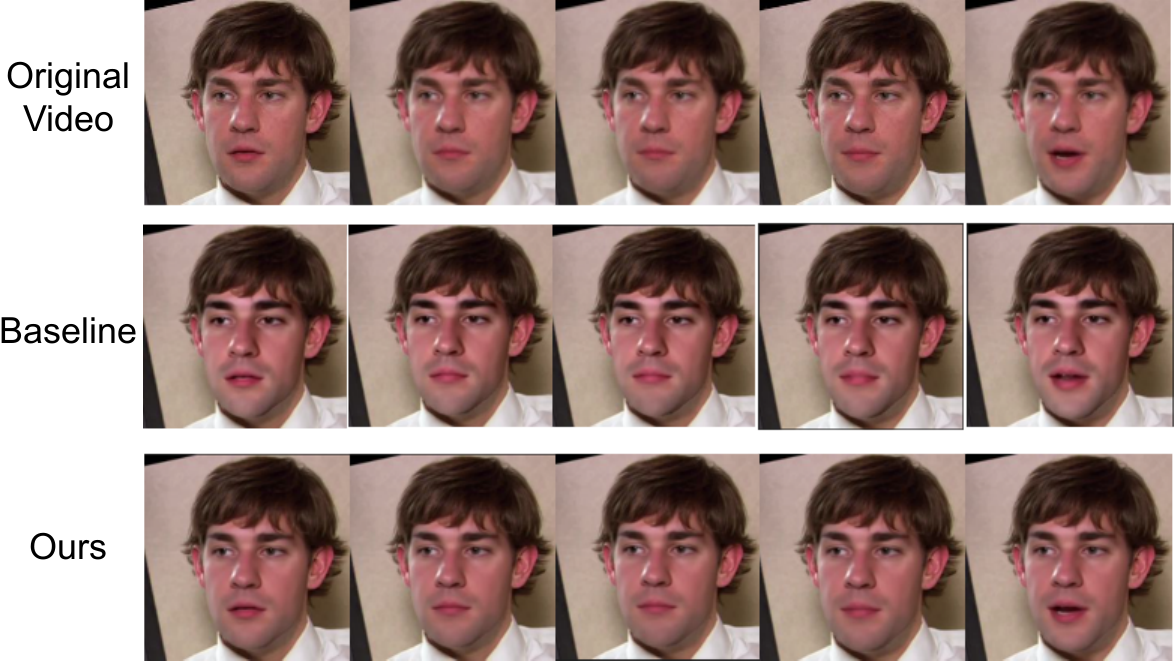}
	\caption{Compared to the baseline method~\cite{kim2023diffusion_videoae}, \algacro{} not only successfully finishes the editing requirement, but also shows better locality for the edit +\texttt{bushy\_eyebrows}.}\label{fig:locality}
\end{figure}

\noindent\textbf{Avoid Over-Editing.}
Figure~\ref{fig:locality} shows that with the edit \texttt{bushy\_eyebrows}, the baseline not only manipulates the eyebrows but also makes the face more hairy, which is unexpected. In contrast, our method not only performed the editing but also kept non-eyebrow areas unchanged, showing better ability to avoid over-editing than the baseline method.

\section{Conclusion}
We propose \algacro{} to improve face video editing methods from the following three perspectives. \textit{(i)} We propose a self-training strategy to improve the generalizability of existing face video editing methods,  which generates pseudo-data in the latent space to further tune a trained face editing model. \textit{(ii)} We design a semantic disentangled editing architecture to help the model accommodate diverse editing requirements. \textit{(iii)} We introduce a sparse learning strategy to facilitate localized editing.
We perform extensive qualitative and quantitative comparisons to show that \algacro{} is compatible with various editing methods, including GAN-based and Diffusion-based methods.  Numerical results validates that \algacro{} consistently improves editing faithfulness, temporal consistency, and identity preservation.

{
    \small
    \bibliographystyle{ieeenat_fullname}
    \bibliography{main}
}


\end{document}